    \documentclass[runningheads]{llncs}
    \usepackage{esvect}

    \usepackage[T1]{fontenc}
    \usepackage{cite}
    \usepackage{amsmath, amssymb, amsfonts}
    \usepackage{algorithmic}
    \usepackage{multirow}
    
    \usepackage{graphicx}
    \usepackage{textcomp}
    \usepackage{xcolor}
    \usepackage{subcaption}
    \usepackage{array}
    \usepackage{booktabs}
    \usepackage{amsmath}
    \usepackage{amssymb}
    \usepackage{booktabs}
    \usepackage{xcolor}
    
    \usepackage{times}
    \usepackage{soul}
    
    \usepackage{multirow}
    \usepackage{array}
    \newcommand{\PreserveBackslash}[1]{\let\temp=\\#1\let\\=\temp}
    \newcolumntype{C}[1]{>{\PreserveBackslash\centering}p{#1}}
    \newcolumntype{R}[1]{>{\PreserveBackslash\raggedleft}p{#1}}
    \newcolumntype{L}[1]{>{\PreserveBackslash\raggedright}p{#1}}
    \usepackage{tikz}

    \usepackage{float}
    \usepackage{mwe}
    
    \graphicspath{{Figures/}}
    \newcommand{\etal}{\textit{et al}.}
    
    \usepackage[pagebackref,breaklinks,colorlinks]{hyperref}
    \usepackage{orcidlink}
    \usepackage{tikz}
    \usepackage[misc]{ifsym}
    \usepackage{bbding} 

    \newcommand{\mobileNet}{\texttt{MobileNetV2\_100 }}
    
    \usepackage{makecell}
    
\begin{document}
    \title{Few-Shot Learning Pipeline for Monkeypox Skin Disease Classification Using CNN Feature Extractors}
    \titlerunning{Few-Shot Learning for Monkeypox Skin Disease Classification}
    %
    %
    \author{(Anonymous BIM Sumbission)}
    \authorrunning{Anonymous et al.}
    \institute{(Anonymous Institution)}
    \author{Md. Safirur Rashid\inst{1}\orcidlink{0009-0003-2444-8192} \and
    Sabbir Ahmed\inst{1}\Envelope~\orcidlink{0000-0001-5928-4886} \and
    Muhammad Usama Islam\inst{1,2}\orcidlink{0000-0003-2080-2484} \and
    Sumona Hoque Mumu\inst{3}\orcidlink{0000-0003-0386-8991} \and
    Md. Hasanul Kabir\inst{1}\Envelope~\orcidlink{0000-0002-6853-8785}}
    \authorrunning{S. Rashid et al.}
    %
    \institute{Department of Computer Science and Engineering, Islamic University of Technology, Gazipur, Bangladesh\\
    \email{\{safirurrashid, sabbirahmed, usamaislam, hasanul\}@iut-dhaka.edu}
    \and
    Management Information Systems, Metropolitan State University, Minnesota, USA
    \\
    \email{muhammadusama.islam@metrostate.edu}
    \and
    School of Kinesiology, University of Louisiana at Lafayette, LA, USA\\
    \email{sumona-hoque.mumu1@louisiana.edu}
    }
    
    %
    \maketitle              
    \begin{abstract}
    Despite the strong performance of Convolutional Neural Networks (CNNs) in disease classification, their effectiveness often depends on access to large annotated datasets, which is an impractical requirement for emerging or rare conditions such as Monkeypox. To overcome this limitation, we propose a few-shot learning (FSL) framework that employs SimpleShot, a lightweight, non-parametric, inductive classifier, for Monkeypox and pox-like skin disease recognition from limited labeled examples. The proposed pipeline passes the skin lesion images through a frozen, pretrained CNN backbone to obtain feature embeddings, which are then classified via SimpleShot using nearest-centroid comparisons in a normalized embedding space. We systematically benchmark six widely used CNN backbones as feature extractors under consistent experimental settings, enabling fair comparison. Experiments on three publicly available datasets (\texttt{MSLD v1.0}, \texttt{MSID}, and \texttt{MSLD v2.0}) are conducted across 2-way, 4-way, and 6-way tasks with 1-shot, 5-shot, and 10-shot configurations. Among all models, \texttt{MobileNetV2\_100} consistently achieves the highest accuracy. In addition, we present a cross-dataset evaluation for Monkeypox classification, revealing that binary Mpox-vs-Others transfer remains comparatively stable while multi-class performance degrades significantly under domain shift. Together, these results demonstrate the practical utility of combining inductive FSL methods with lightweight CNN backbones and highlight the importance of domain robustness for reliable real-world clinical deployment.
    \end{abstract}
    

    
    \section{Introduction}
    
    Monkeypox is a re-emerging zoonotic disease caused by the Monkeypox virus (MPXV), a member of the \textit{Orthopoxvirus} genus \cite{Mitjà2022Monkeypox}. 
    It clinically manifests with skin lesions that are visually similar to other pox-like illnesses such as Chickenpox, Measles, and Cowpox, making accurate clinical diagnosis particularly challenging in early stages or in low-resource settings \cite{Shafaati2022State-of-the-art, Al-Musa2022The}. With the global resurgence of Monkeypox cases especially in non-endemic regions there is an increasing need for automated diagnostic systems that can aid in triage and early detection \cite{Quarleri2022Monkeypox:, Nisar2023A, McCollum2023Epidemiology, Parums2022Editorial:}.

    The automatic feature learning capability of modern deep learning models has revolutionized the field of artificial intelligence, driving state-of-the-art solutions across diverse downstream tasks in computer vision \cite{hasan2023GaitGCN, ahmed2024exeNet, ivan2024meme, raiyan2025hasper,fuad2025aqua20}, speech recognition \cite{aziz2023banglaSER,yasmeen2021csvcNet}, natural language processing \cite{ahmed2024Depression,ridwan2023poem}, and related domains\cite{rahman2022twoDecades,ahmed2022less,khan2022rethinking,arif2025hallucinaton}. Within medical data analysis in particular, these advances have enabled automated systems to achieve expert-level performance in tasks once considered uniquely dependent on clinical expertise \cite{ahmed2023structure,khan-etal-2023-banglachq, ajwad2025banglaCHQ, alvi2023nervous,farzana2024cancer}. Deep learning (DL), especially convolutional neural networks (CNNs), now serves as the backbone of medical image understanding, powering systems for dermatology, radiology, and histopathology \cite{Cai2020Few-Shot,Jiang2022Multi-Learner}. In skin disease classification, CNNs trained on large-scale datasets such as ISIC \cite{Cassidy2021Analysis} have already reached dermatologist-level accuracy \cite{khan2023attResDUNet}. 
    Motivated by this success, recent works have explored applying DL to Monkeypox detection \cite{BALA2023757,Deng2024Using,Haque2022Classification,Kottath2025PoxTLNet50DL}. For instance, MonkeyNet \cite{BALA2023757} leverages a DenseNet backbone to discriminate pox-like skin conditions with high accuracy, while others integrate attention mechanisms \cite{Haque2022Classification} or self-supervised learning strategies \cite{Deng2024Using}. More recently, federated learning approaches have been introduced to mitigate data-sharing and privacy challenges \cite{Kundu2024Federated}.
    
    Despite promising results, most existing approaches primarily adopt conventional supervised learning pipelines, often supplemented with data augmentation strategies to artificially expand the training set \cite{Islam2023Medical}. While such methods can provide marginal improvements, they remain fundamentally constrained by the availability and diversity of labeled data. In the case of Monkeypox, datasets are still relatively small, highly imbalanced, and difficult to curate due to the challenges of obtaining clinically verified images across different stages of infection, varying skin tones, and real-world conditions. This scarcity not only limits the generalizability of supervised models but also amplifies risks of overfitting, where models may inadvertently exploit spurious correlations instead of learning robust disease-relevant features \cite{ali2022msldv1,ali2023msldv2}. These limitations highlight the need for alternative paradigms, such as few-shot learning, aiming to achieve strong generalization under low-data regimes by leveraging transferable representations.
    
    
    Few-shot learning (FSL) has emerged as a powerful paradigm for addressing extreme data scarcity by enabling models to generalize from only a handful of labeled examples per class \cite{ahmed2025dexnet, mehedi2024fslBHDR}. Comprehensive surveys have chronicled its evolution from early metric-learning approaches to meta-learning and self-supervised methods, emphasizing its growing role in low-resource machine learning scenarios \cite{Wang2022A, Wang2019Few-shot}. In the medical domain, FSL has shown considerable success in tasks such as disease classification \cite{SINGH2021108111}, hyperspectral imaging \cite{Liu2019Deep}, and medical image segmentation \cite{Ouyang2022Self-Supervised}, retinal imaging \cite{SINGH2021108111}, cancer detection \cite{Dai2023PFEMed:}, and multimodal classification \cite{Zhang2023A} reducing the need for large annotated datasets while retaining strong generalization capabilities. However, FSL remains largely underexplored in the context of Monkeypox and other pox-like skin diseases. While some recent works have introduced tailored few-shot pipelines that incorporate domain adaptation via pretrained encoder fine-tuning \cite{Chen2023A}, these approaches often rely on access to auxiliary datasets. 
    
    In contrast, our work explores a purer inductive few-shot learning setup, eschewing domain adaptation entirely by pairing frozen convolutional feature extractors with a non-parametric classifier to assess generalization under low-data constraints. Our proposed FSL pipeline is tailored for Monkeypox and pox-like skin disease classification, centered around the SimpleShot algorithm  \cite{wang2019simpleshot}, a robust prototype-based algorithm that performs nearest-centroid classification in a normalized embedding space. We adopt the SimpleShot algorithm not only for its simplicity and efficiency, but also because it achieves strong empirical performance on standard benchmarks without requiring costly meta-training, making it ideal for low-resource clinical settings.
    
    To the best of our knowledge, this is the first systematic evaluation of CNN-based feature extractors for Monkeypox classification using a standardized few-shot pipeline. Our contribution lies in establishing a reproducible framework, centered around the SimpleShot algorithm, in which we benchmark six widely used CNN backbones as frozen feature encoders. Unlike prior efforts focusing on either novel FSL algorithms or fully supervised deep learning approaches, we shift focus toward understanding which CNN backbones are most effective for low-data skin disease classification, and how this choice influences model generalization, efficiency, and clinical applicability.
    
    To evaluate our SimpleShot-centered pipeline, we benchmark six widely used pretrained CNNs, \texttt{VGG16} \cite{simonyan2015vgg16}, \texttt{InceptionV3} \cite{szegedy2015inceptionv3}, \texttt{ResNet50} \cite{he2015resnet50}, \texttt{DenseNet121} \cite{huang2018densenet121}, \texttt{MobileNetV2\_100} \cite{sandler2019mobilenetv2}, and \texttt{EfficientNet\_B1} \cite{tan2020efficientnet} as frozen feature extractors, feeding their output embeddings directly into the SimpleShot classifier to isolate the impact of backbone choice. Experiments are conducted on three publicly available skin lesion datasets, \texttt{MSLD v1.0} \cite{ali2022msldv1}, \texttt{MSID} \cite{BALA2023757}, and \texttt{MSLD v2.0} \cite{ali2023msldv2} across 2-way, 4-way, and 6-way classification tasks under 1-shot, 5-shot, and 10-shot configurations. Each setup is evaluated over 100 few-shot episodes with 50 query samples, and performance is reported as mean accuracy with 95\% confidence intervals. Results show that \mobileNet achieves the best balance between accuracy and computational efficiency, while increasing support size consistently improves performance. Class-level analysis reveals higher confusion among visually similar conditions, such as Monkeypox and Chickenpox, reinforcing the need for strong feature encoders in low-data clinical environments. 
    The key contributions of this work are as follows:
    \begin{itemize}
        \item We introduce a simple, scalable, and effective few-shot learning pipeline for Monkeypox skin disease classification, built entirely around the SimpleShot algorithm.
        \item We benchmark six widely used CNN backbones as feature extractors within this pipeline and identify \mobileNet as the most effective architecture.
        \item Upon benchmarking on three datasets, we analyze the impact of class count, support size, and disease similarity on classification accuracy, providing practical insights for deploying FSL models in real-world clinical screening scenarios.
        \item We conduct a comprehensive cross-dataset evaluation for Monkeypox skin disease classification, showing that binary Mpox-vs-Others transfer remains relatively stable (63–68\%), whereas multi-class generalization degrades significantly under domain shift, highlighting the need for domain-adaptive FSL strategies to ensure reliable clinical deployment across diverse data sources. 
    
    \end{itemize}

    \section{Methodology}
\begin{figure}[h!]
        \centering
        \includegraphics[width=.95\linewidth]{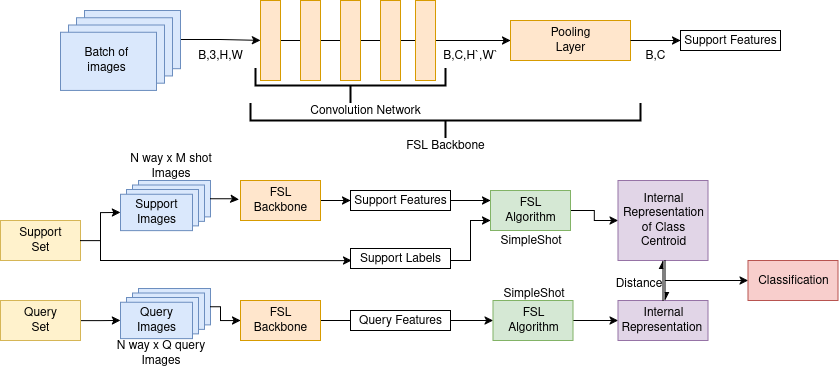}
        \caption{Overview of the proposed pipeline: input images are processed by a frozen CNN backbone to extract feature embeddings, which are then classified using SimpleShot in an $N$-way $M$-shot setting.}
        \label{fig:pipeline_diagram}
    \end{figure}

    This study proposes a streamlined and reproducible FSL pipeline for Monkeypox and pox-like skin disease classification, with a specific focus on identifying the most effective convolutional neural network (CNN) backbone for feature extraction with limited data. At the core of our pipeline lies SimpleShot \cite{wang2019simpleshot}, a lightweight and efficient prototype-based FSL algorithm that classifies query samples by computing distances to class centroids in a normalized embedding space. We adopt SimpleShot not only for its speed and simplicity, but also because it eliminates the need for costly meta-training and enables direct benchmarking of backbone quality. The overall experimental workflow is illustrated in Figure~\ref{fig:pipeline_diagram}, showing the pipeline from image input through CNN-based embedding extraction and final classification via SimpleShot.

    
    \vspace{-1cm}
    \subsection{Dataset Overview}
    \label{dataset}
    We utilized three publicly available skin lesion datasets, MSLD v1.0 \cite{ali2022msldv1}, MSID \cite{BALA2023757}, and MSLD v2.0 \cite{ali2023msldv2} that differ in size, class diversity, and visual complexity. All experiments are conducted exclusively on the original, unaugmented versions to avoid bias and ensure consistency in backbone evaluation.
    We evaluate model performance across 2-way, 4-way, and 6-way classification scenarios using $M$-shot settings with $M \in \{1, 5, 10\}$ and a fixed query set of 50 samples per episode. Each configuration is repeated over 100 randomly sampled tasks to ensure statistical robustness. Results are reported as mean accuracy along with 95\% confidence intervals.

\begin{figure}[H]
        \centering
    
        \begin{subfigure}[b]{0.25\textwidth}
            \includegraphics[width=\linewidth]{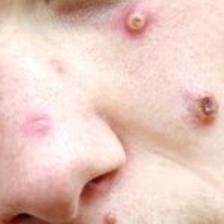}
            \caption{Monkeypox}
        \end{subfigure}
        \hspace{0.05\textwidth}
        \begin{subfigure}[b]{0.25\textwidth}
            \includegraphics[width=\linewidth]{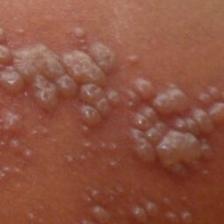}
            \caption{Others}
        \end{subfigure}
    
        \caption{Example images from the MSLD v1.0 dataset \cite{ali2022msldv1}. (a) Monkeypox: deep-seated pustules with defined borders and occasional central umbilication. (b) Chickenpox: clusters of small vesicles on an erythematous base, more superficial and variable in size, sometimes resembling shingles. Such features can overlap with early Monkeypox, thereby increasing diagnostic complexity.}
        \label{fig:msldv1_samples}
    \end{figure}

    
    \textit{\textbf{Monkeypox Skin Lesion Dataset v1.0 (MSLD v1.0)} 
    }: MSLD v1.0, released during the early stages of the 2022 Monkeypox outbreak~\cite{ali2022msldv1}, is a binary classification dataset comprising 228 clinical images: 102 of Monkeypox and 126 from a merged `Others' class that includes Chickenpox and Measles. Figure~\ref{fig:msldv1_samples} shows a sample from each class, illustrating the diagnostic challenge posed by the visual overlap between Monkeypox and other viral skin diseases.
    
    \vspace{-.5cm}

    Images were sourced from publicly available repositories and subjected to a two-stage quality control process, involving duplicate removal and clinical verification via manual screening and Google Reverse Image Search. Although subsequent studies have expanded this dataset with augmented samples, our work confines evaluation strictly to the original collection to prevent potential data leakage. Despite its modest scale, MSLD v1.0 serves as a valuable low-shot, 2-way classification benchmark. The heterogeneous composition of the \textit{Others} class introduces significant intra-class variability, simulating diagnostic ambiguity and providing a challenging testbed for early screening scenarios.

    \textit{\textbf{Monkeypox Skin Lesion Dataset (MSLD v2.0)}}:
    MSLD v2.0 is a six-class dataset comprising 755 high-quality clinical and dermoscopic skin images from 541 unique patients. The classes include Monkeypox (284), Chickenpox (75), Measles (55), Cowpox (66), Hand-Foot-Mouth Disease (161), and Healthy skin (114), collectively covering a broad spectrum of visually overlapping conditions relevant for differential diagnosis.
    
\begin{figure}[t]
        \centering
    
        \begin{subfigure}[b]{0.22\textwidth}
            \includegraphics[width=\linewidth]{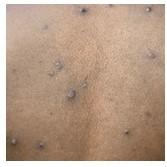}
            \caption{Monkeypox}
        \end{subfigure}
        \hspace{0.05\textwidth}
        \begin{subfigure}[b]{0.22\textwidth}
            \includegraphics[width=\linewidth]{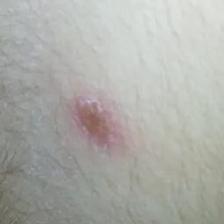}
            \caption{Chickenpox}
        \end{subfigure}
        \hspace{0.05\textwidth}
        \begin{subfigure}[b]{0.22\textwidth}
            \includegraphics[width=\linewidth]{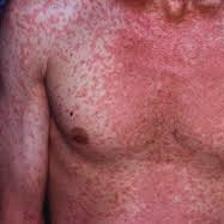}
            \caption{Measles}
        \end{subfigure}
    
        \vspace{1em}
    
        \begin{subfigure}[b]{0.22\textwidth}
            \includegraphics[width=\linewidth]{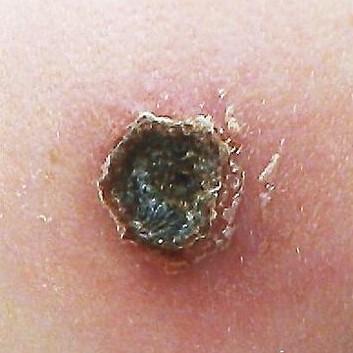}
            \caption{Cowpox}
        \end{subfigure}
        \hspace{0.05\textwidth}
        \begin{subfigure}[b]{0.22\textwidth}
            \includegraphics[width=\linewidth]{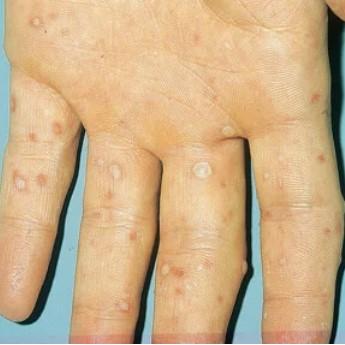}
            \caption{HFMD}
        \end{subfigure}
        \hspace{0.05\textwidth}
        \begin{subfigure}[b]{0.22\textwidth}
            \includegraphics[width=\linewidth]{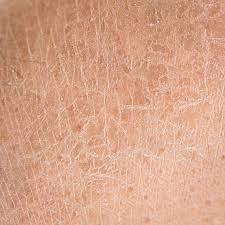}
            \caption{Healthy}
        \end{subfigure}
    
        \caption{Example images from the MSLD v2.0\cite{ali2023msldv2} dataset. (a) Monkeypox: well-circumscribed pustules with central umbilication. (b) Chickenpox: superficial vesicles on red bases at mixed stages. (c) Measles: widespread confluent maculopapular rash. (d) Cowpox: necrotic lesion with central eschar. (e) HFMD: vesicles localized to palms, soles, or mucosa. (f) Healthy: normal skin without lesions, serving as control.}
    
        \label{fig:msldv2_samples}
    \end{figure}

    Developed by Ali \etal~\cite{ali2023msldv2}, MSLD v2.0 extends earlier efforts~\cite{BALA2023757, ali2022msldv1} by significantly enlarging the sample base. All images were collected via web scraping, de-identified to protect patient privacy, and clinically validated by infectious disease experts. Although the full dataset includes over 39,000 augmented samples, our experiments are conducted exclusively on the original 755 images. Representative examples are shown in Figure~\ref{fig:msldv2_samples}, highlighting the strong visual similarity among conditions such as Monkeypox, Chickenpox, and Cowpox.
    
    Overall, MSLD v2.0 presents a challenging 6-way classification benchmark characterized by class imbalance and substantial inter-class ambiguity, particularly among pox-like diseases. Its clinical diversity and diagnostic complexity make it a valuable testbed for assessing backbone generalization and robustness in FSL frameworks.
    
\begin{figure}[h!]
        \centering
    
        \begin{subfigure}[b]{0.22\textwidth}
            \includegraphics[width=\linewidth]{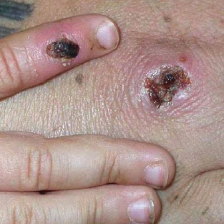}
            \caption{Monkeypox}
        \end{subfigure}
        \hfill
        \begin{subfigure}[b]{0.22\textwidth}
            \includegraphics[width=\linewidth]{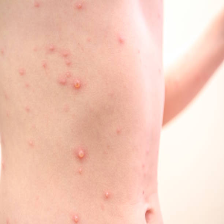}
            \caption{Chickenpox}
        \end{subfigure}
        \hfill
        \begin{subfigure}[b]{0.22\textwidth}
            \includegraphics[width=\linewidth]{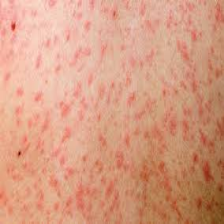}
            \caption{Measles}
        \end{subfigure}
        \hfill
        \begin{subfigure}[b]{0.22\textwidth}
            \includegraphics[width=\linewidth]{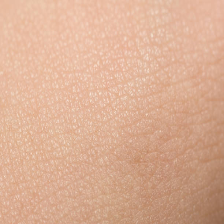}
            \caption{Healthy}
        \end{subfigure}
    
        \caption{Samples from the MSID\cite{BALA2023757} dataset. (a) Monkeypox: firm pustules with umbilication and surrounding erythema. (b) Chickenpox: fragile vesicles at different stages. (c) Measles: diffuse erythematous maculopapular rash. (d) Healthy: skin without visible lesions or inflammation.}
    
        \label{fig:msid_samples}
    \end{figure}

    \textit{\textbf{Monkeypox Skin Images Dataset (MSID)}}:
    This dataset was curated by Bala \etal~\cite{BALA2023757}, consisting of 770 clinical images grouped into four diagnostic classes: Monkeypox (279), Chickenpox (107), Measles (91), and Healthy (293). The samples were sourced from medical literature, dermatological atlases, and publicly available image repositories, with a focus on diseases commonly misdiagnosed due to visual similarity. Representative class-wise samples are presented in Figure~\ref{fig:msid_samples}, highlighting both inter-class similarities and visual variance in skin tone, lighting, and lesion type.
    
    \vspace{-.5cm}

    Although MSID was later expanded to over 8,600 samples through aggressive augmentation, our experiments rely solely on the original, unaugmented images. The dataset features images from various anatomical regions and lighting conditions, adding to its robustness and clinical realism. MSID is positioned as an intermediate-complexity dataset for 4-way classification, bridging the binary structure of MSLD v1.0 and the more complex MSLD v2.0.

    \textit{\textbf{Dataset Limitations}}: 
    The datasets suffer from inherent class imbalance and limited coverage of skin tone diversity, both of which constrain model performance on underrepresented categories. In particular, MSLD v1.0 and MSID contain only a small number of samples for certain diseases, while MSLD v2.0, despite being larger, still exhibits uneven distribution across conditions. Furthermore, as all datasets are primarily sourced from publicly available web repositories, they lack racial and geographic diversity. These limitations not only reduce the reliability of accuracy estimates for minority classes but also raise concerns about the robustness and fairness of model generalization to real-world clinical populations.
    
    \subsection{Preprocessing}
    To ensure architectural compatibility and maintain experimental consistency, all input images were resized to $128 \times 128$ for consistent feature extraction. No additional preprocessing or data augmentation techniques were applied. This deliberate choice was made to isolate the true representation capability of each CNN backbone, without confounding the results with synthetic variability introduced through augmentation. By maintaining the original (resized-only) image content, we ensure a fair and controlled comparison of feature extraction quality across models. Each resized image was directly passed through the corresponding backbone network to extract deep feature embeddings, which were subsequently used as input to the downstream few-shot classifier.
    
    \subsection{Feature Extraction via CNN Backbones}
    Our pipeline begins by extracting feature embeddings from input images using one of six widely adopted CNN backbones: {VGG16}, {InceptionV3}, {ResNet50}, {DenseNet121}, MobileNetV2\_100, and {EfficientNet B1}. Each model is initialized with ImageNet pretrained weights and kept entirely frozen during inference to ensure a fair and unbiased comparison across architectures. Feature maps are taken from the final convolutional layer of each network, producing a tensor of shape $(B, C, H', W')$ for a batch of $B$ images. We then apply a \texttt{SelectAdaptivePool2d} operation, which compresses each spatial feature map to shape $(B, C)$ as shown in \equationautorefname~\ref{eq:adaptive_pooling} and passed directly to the SimpleShot classifier.
    
    Let $\mathbf{F} \in \mathbb{R}^{B \times C \times H' \times W'}$ be the feature maps obtained from the last convolutional layer of a CNN. The adaptive pooling operation $\mathcal{P}$ reduces  spatial dimensions such that:
        \begin{equation}
        \mathcal{P}(\mathbf{F}) = \frac{1}{H' W'} \sum_{i=1}^{H'} \sum_{j=1}^{W'} \mathbf{F}_{:, :, i, j} \in \mathbb{R}^{B \times C \times 1 \times 1}
        \label{eq:adaptive_pooling}
        \end{equation}
        
    This pooled tensor is then reshaped to $\mathbb{R}^{B \times C}$ to obtain a $(B, C)$ feature matrix before being passed to the SimpleShot classifier.
        
    This process ensured that all embeddings, regardless of backbone architecture, had consistent structure and were spatially invariant, allowing a fair comparison of their representational power for few-shot classification.
        
    \subsection{Few-shot Classification via SimpleShot}
    For all classification tasks, we employed SimpleShot~\cite{wang2019simpleshot}, a non-parametric and inductive few-shot learning algorithm that requires no meta-training. SimpleShot operates by computing class prototypes from the support set and classifying query samples via nearest-centroid matching in a normalized embedding space. Its simplicity, efficiency, and strong baseline performance make it an ideal choice for evaluating the representational power of CNN-based feature extractors without introducing algorithm-specific learning biases. Furthermore, follow-up studies have shown that SimpleShot is the best-performing non-parametric and inductive FSL method in challenging cross-domain few-shot settings~\cite{sekhar2024cross}, often outperforming more complex meta-learning approaches while maintaining minimal computational overhead. This makes it particularly well-suited as a standardized backbone evaluation tool in our low-data clinical scenario.
    
    Each experiment follows an $N$-way $M$-shot protocol, where $N \in \{2, 4, 6\}$ varies by dataset and $M \in \{1, 5, 10\}$ defines the number of support examples per class. For each classification episode, a support set is randomly sampled with $M$ labeled examples per class, while a fixed query set of 50 images is used for evaluation. Every configuration is repeated over 100 randomly sampled episodes, and mean accuracy with 95\% confidence intervals is reported to ensure statistical robustness.
    
    Let $\mathbf{z}_i \in \mathbb{R}^C$ denote the embedding of a support sample and let $\mathcal{S}_c$ be the set of support embeddings for class $c$. The prototype for class $c$ is computed as the mean of its support embeddings:
    \begin{equation}
    \mathbf{p}_c = \frac{1}{|\mathcal{S}_c|} \sum_{\mathbf{z}_i \in \mathcal{S}_c} \mathbf{z}_i,
    \label{eq:prototype}
    \end{equation}
    where $\mathbf{p}_c \in \mathbb{R}^C$ is the class prototype. During inference, query samples are classified to the nearest prototype we get from \equationautorefname~\ref{eq:prototype} in the normalized embedding space. This formulation ensures consistent evaluation across varying task complexities and support sizes, enabling systematic assessment of feature extractor generalizability under different levels of class granularity and data scarcity.
    
    \subsection{Evaluation Metrics}
    Model performance was evaluated using the mean classification accuracy across 100 independently sampled few-shot classification tasks for each configuration. All experiments were conducted using Google Colab on a single NVIDIA T4 GPU. 
    To quantify the statistical reliability of each result, we computed 95\% confidence intervals using the \texttt{mean\_confidence\_interval} function, which estimates the interval based on the Student's t-distribution. Formally, each result is reported as \( \mu \pm t_{n-1} \cdot \frac{\sigma}{\sqrt{n}} \); where $\mu$ is the sample mean accuracy, $\sigma$ is the sample standard deviation, $n = 100$ is the number of episodes, and $t_{n-1}$ is the critical value of the t-distribution with $n - 1$ degrees of freedom at 95\% confidence.
    The reporting format captures both central tendency and variability across episodes, enabling robust comparisons across backbones, datasets, and shot configurations. 
        
        
    \vspace{-2mm}
    \section{Results and Discussion}
    \vspace{-2mm}
    This section presents our experimental results and insights, where we begin by benchmarking multiple CNN backbones across datasets of increasing complexity. Next, we quantify the impact of increasing support set size ($N$-shot) on classification accuracy. Furthermore, we analyze the class-wise performance using the best-performing model and provide a thorough analysis of cross-dataset generalization.
    
    \vspace{-2mm}
    \subsection{Backbone Comparison Across Datasets}
    \label{sec:backbone_results}
    
    \tableautorefname~\ref{tab:fsl-results} reports the average classification accuracy (with 95\% confidence intervals) for six CNN models across three datasets under 1-shot, 5-shot, and 10-shot conditions.

    \vspace{-.5cm}
    \begin{table}[h]
\centering
\caption{Performance analysis of proposed pipeline with different CNN feature
extractors accross different datasets.}
\label{tab:fsl-results}
\resizebox{\textwidth}{!}{%
\begin{tabular}{
L{0.18\textwidth}
C{0.1\textwidth}C{0.1\textwidth}C{0.1\textwidth}
C{0.1\textwidth}C{0.1\textwidth}C{0.1\textwidth}
C{0.1\textwidth}C{0.1\textwidth}C{0.1\textwidth}
}
\hline
\textbf{Backbone} & 
\multicolumn{3}{c}{\textbf{6-way (\texttt{MSLD v2.0})}} & 
\multicolumn{3}{c}{\textbf{4-way (\texttt{MSID})}} & 
\multicolumn{3}{c}{\textbf{2-way (\texttt{MSLD v1.0})}} \\

\cmidrule(lr){2-4} \cmidrule(lr){5-7} \cmidrule(lr){8-10}& 
\textbf{10-shot} & \textbf{5-shot} & \textbf{1-shot} & 
\textbf{10-shot} & \textbf{5-shot} & \textbf{1-shot} & 
\textbf{10-shot} & \textbf{5-shot} & \textbf{1-shot} \\

\hline
\texttt{MobileNet}\newline \texttt{v2\_100} &
\boldmath$0.624\pm0.006$ & \boldmath$ 0.539\pm0.007$ & $0.351\pm0.010$ & 
\boldmath$0.696\pm0.008$ & \boldmath$0.609\pm0.011$ & \boldmath$0.443\pm0.015$ & 
$0.674\pm0.010$ & $0.642\pm0.013$ & $0.548\pm0.017$ \\

\texttt{VGG16} & 
$0.566\pm0.006$ & $0.516\pm0.008$ & \boldmath$0.358\pm0.011$ & 
$0.641\pm0.008$ & $0.579\pm0.010$ & $0.440\pm0.014$ & 
\boldmath$0.710\pm0.011$ & \boldmath$0.676\pm0.011$ & \boldmath$0.599\pm0.018$ \\

\texttt{Efficient}\newline\texttt{Net\_B1} & 
$0.537\pm0.007$ & $0.459\pm0.008$ & $0.296\pm0.009$ & 
$0.632\pm0.008$ & $0.553\pm0.009$ & $0.411\pm0.014$ & 
$0.644\pm0.009$ & $0.618\pm0.012$ & $0.567\pm0.017$ \\

\texttt{DenseNet121} & 
$0.523\pm0.006$ & $0.450\pm0.009$ & $0.305\pm0.009$ & 
$0.606\pm0.009$ & $0.523\pm0.010$ & $0.389\pm0.014$ & 
$0.634\pm0.012$ & $0.604\pm0.014$ & $0.531\pm0.015$ \\

\texttt{ResNet50} & 
$0.498\pm0.007$ & $0.431\pm0.010$ & $0.288\pm0.010$ & 
$0.548\pm0.009$ & $0.475\pm0.010$ & $0.359\pm0.012$ & 
$0.643\pm0.011$ & $0.598\pm0.013$ & $0.533\pm0.015$ \\

\texttt{InceptionV3} & 
$0.336\pm0.006$ & $0.288\pm0.007$ & $0.213\pm0.007$ & 
$0.397\pm0.008$ & $0.344\pm0.007$ & $0.282\pm0.007$ & 
$0.578\pm0.010$ & $0.556\pm0.011$ & $0.496\pm0.011$ \\
\bottomrule
\end{tabular}
}
\end{table}

    \vspace{-.3cm}


    Across nearly all experimental configurations, \mobileNet consistently outperforms other CNN backbones, underscoring its strong generalization capability and making it particularly well-suited for low-data regimes common in few-shot learning scenarios. We observe a clear positive correlation between the number of support examples and classification accuracy, reaffirming that increasing the number of labeled instances enhances the model’s ability to learn discriminative features. However, task complexity plays a significant role in performance degradation as the classification challenge shifts from 2-way to 6-way settings, accuracy drops considerably, likely due to increased inter-class visual similarity and ambiguity. Interestingly, larger or deeper models such as InceptionV3 and VGG16 do not necessarily yield superior results, often underperforming compared to more compact architectures. This suggests that lightweight models with effective inductive biases may be more advantageous than computationally heavier counterparts in the context of few-shot learning.
    
    To substantiate these findings, we also report inference time, parameter count, and floating point operations (FLOPs) for each backbone in \tableautorefname~\ref{tab:efficiency-metrics}. ~\mobileNet achieves the lowest latency and computational cost while maintaining superior accuracy, reinforcing its suitability for deployment in resource-constrained environments.
    
    
    \begin{table}[t]
    \centering
    \caption{Efficiency metrics of different backbone models evaluated on a single NVIDIA T4 GPU. Parameters are in millions (M), FLOPs in millions (M), and inference time per image in milliseconds (ms/img).}
    \label{tab:efficiency-metrics}
    \resizebox{0.85\linewidth}{!}{%
    \begin{tabular}{p{3.2cm} p{2.5cm} p{2cm} l} 
    \toprule
    Backbone & Parameters (M) &  FLOPs (M) & Inference Time (ms/img) \\
    \midrule
    \texttt{MobileNetV2\_100}  & 1.81  & 186.42   & 2.84  \\
    \texttt{VGG16}              & 14.71 & 10,060   & 7.66  \\
    \texttt{EfficientNet\_B1}  & 6.10  & 365.04   & 15.32 \\
    \texttt{DenseNet121}        & 6.95  & 1,872.64 & 19.33 \\
    \texttt{ResNet50}           & 23.51 & 2,700    & 7.38  \\
    \texttt{InceptionV3}        & 21.79 & 1,468.72 & 15.44 \\
    \bottomrule
    \end{tabular}
    }
    \end{table}
    \vspace{-.3cm}


    
    

    \subsection{Impact of N-Shot Scaling}
\begin{figure}[h!]
        \centering
        \includegraphics[width=0.64\linewidth]{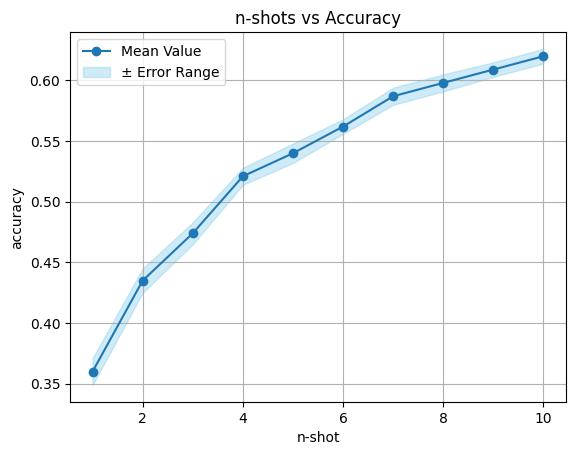}
        \caption{Accuracy improvement with increasing support size  on the 6-way classification task (Error bars represent 95\% confidence intervals). }
        \label{fig:nshot-scaling}
    \end{figure}

    \vspace{-.1cm}
    \label{sec:nshot_scaling}
    To further investigate the influence of support set size, we conducted a 6-way classification experiment using \mobileNet. Figure~\ref{fig:nshot-scaling} illustrates the accuracy trend as $N$ increases from 1 to 10. We observe a significant performance gain between 1-shot and 5-shot settings, with a more gradual improvement from 5 to 10 shots. These results are consistent with existing FSL literature, wherein larger support sets facilitate more robust decision boundaries by reducing intra-class variance.

    \subsection{Per-Class Classification Performance}
    \label{sec:classwise_perf}
    
\begin{figure}[ht]
        \centering
        \includegraphics[width=\linewidth]{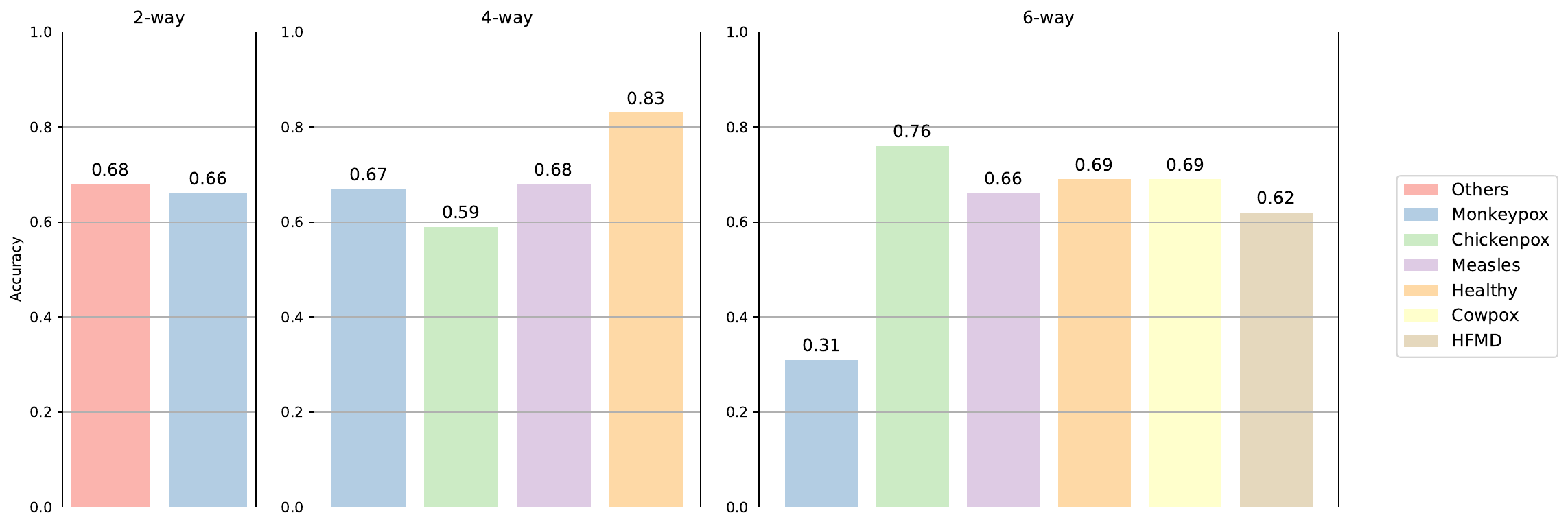}
        \caption{Class-wise accuracy using the \texttt{MobileNetV2\_100} backbone across 2-way, 4-way, and 6-way tasks. Accuracy for Monkeypox decreases as the number of classes increases, indicating greater inter-class confusion in more complex settings.}
    
        \label{fig:classwise-accuracy}
    \end{figure}

\begin{figure}[h!]
        \centering
        \includegraphics[width=0.65\linewidth]{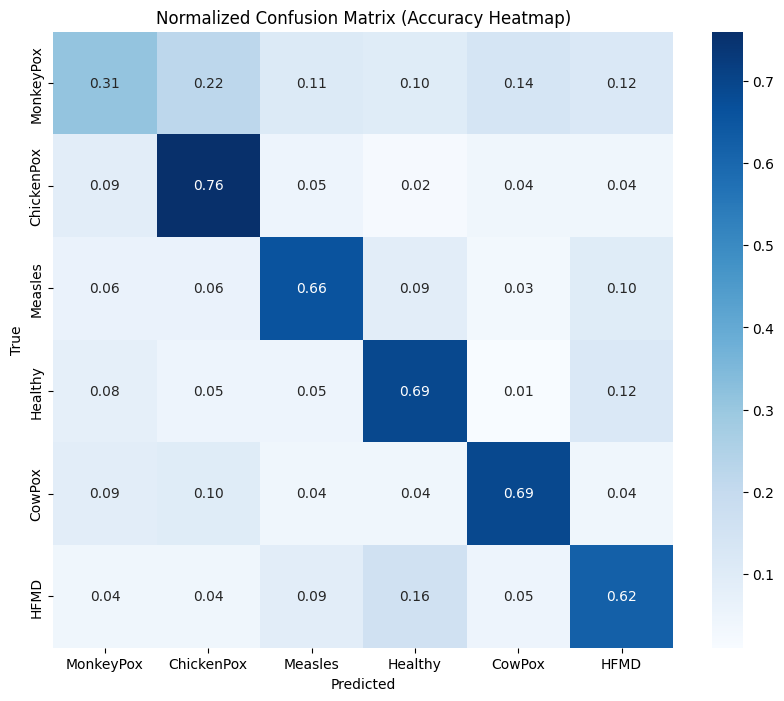}
        \caption{Confusion matrix on the 6-way MSLD v2.0 task (10-shot). Misclassifications are concentrated among visually similar diseases, highlighting inter-class confusion.}
        \label{fig:confusion-matrix}
    \end{figure}

    We assessed class-wise accuracy using \mobileNet under 2-way, 4-way, and 6-way settings (Figure~\ref{fig:classwise-accuracy}). While other classes consistently achieve high accuracy, Monkeypox performance drops notably in the 6-way task. To investigate this decline, we analyzed the confusion matrix (Figure~\ref{fig:confusion-matrix}), which reveals substantial misclassifications between Monkeypox, Chickenpox, Cowpox, and HFMD. This pattern reflects the strong visual overlap among these conditions, where lesions share similar morphology and distribution. These findings highlight the inherent difficulty of distinguishing visually related diseases in low-data diagnostic settings.
    
    
    
    \vspace{-1.2cm}
    
    \subsection{Cross-Dataset Generalization}
    \label{sec:cross_dataset}
    
    We conducted cross-dataset experiments between MSLD v2.0 and MSID to evaluate robustness under domain shift, where prototypes were constructed using one dataset (support set) and classification was performed on the other (query set). Results are summarized in Table~\ref{tab:cross-dataset}.
    
    While using the full 6-class MSLD v2.0 support set to classify MSID queries, accuracy dropped sharply to around $44\%$, compared to in-domain results above $60\%$, highlighting the severity of domain shift. Restricting evaluation to the four overlapping classes yielded much more stable results, with performance converging near $57\%$–$58\%$ regardless of which dataset provided the support set. In the binary Mpox-vs-Others setting, cross-dataset transfer was more reliable, reaching about $63\%$–$68\%$ accuracy depending on the direction. 
    These findings show that while cross-dataset generalization is feasible, domain differences substantially affect multi-class performance. The results further suggest that simpler binary diagnostic setups are more robust, while higher-way classification remains vulnerable to inter-dataset discrepancies.
    
    \vspace{-.6cm}
    
    \begin{table}[h]
    \centering
    \caption{Cross-dataset few-shot classification accuracy (mean $\pm$ 95\% CI). Support and query sets are drawn from different datasets to assess generalization.}
    \label{tab:cross-dataset}
    \resizebox{.9\textwidth}{!}{%
    \begin{tabular}{p{0.3\textwidth} p{0.25\textwidth} p{0.25\textwidth}p{0.15\textwidth}}
    \toprule
    \textbf{Experiment Type} & \textbf{Support Set} & \textbf{Query Set} & \textbf{Accuracy} \\
    \midrule
    In-domain (baseline)& MSLD v2.0 (6-way) & MSLD v2.0 (6-way) & \textbf{0.624 $\pm$ 0.006} \\
                                  & MSID (4-way)      & MSID (4-way)      & \textbf{0.696 $\pm$ 0.008} \\
    \midrule
    Cross-dataset (mismatch) & MSLD v2.0 (6-way) & MSID (4-way) & 0.444 $\pm$ 0.009 \\
    \midrule
    \makecell[l]{Cross-dataset\\(4-class overlap)}
    & MSLD v2.0 (4-way) & MSID (4-way) & 0.570 $\pm$ 0.009 \\
                                             & MSID (4-way)      & MSLD v2.0 (4-way) & 0.578 $\pm$ 0.011 \\
    \midrule
    \makecell[l]{Cross-dataset\\(binary Mpox vs Others)}
    & MSLD v2.0 (2-way) & MSID (2-way) & 0.679 $\pm$ 0.016 \\
                                                    & MSID (2-way)      & MSLD v2.0 (2-way) & 0.631 $\pm$ 0.010 \\
    \bottomrule
    \end{tabular}%
    }
    \end{table}
    
    \vspace{-1cm}

    \section{Conclusion}
    
    We evaluate six CNN backbones for few-shot classification of Monkeypox and pox-like skin diseases. Across three datasets and multiple task complexities, \texttt{MobileNetV2\_100} achieves the best balance of accuracy and efficiency. Performance improves substantially from 1-shot to 5-shot settings, with limited gains beyond that. Per-class analysis shows persistent confusion among visually similar diseases, highlighting the importance of backbone selection. Cross-dataset experiments indicate that binary Mpox-vs-Others transfer remains relatively stable (63–68\%), whereas multi-class generalization declines under domain shift, underscoring the need for domain-adaptive strategies. These results demonstrate the practical value of lightweight CNNs with inductive few-shot methods, and future work may explore self-supervised pretraining and domain adaptation to enhance robustness across clinical datasets.


    \bibliography{references}
    
    \end{document}